\documentclass{IOS-Book-Article}

\usepackage{mathptmx}
\usepackage[table,dvipsnames]{xcolor}
\usepackage{graphicx}
\usepackage{relsize}
\usepackage{soul}\setuldepth{article}
\usepackage{url}
\usepackage{amsmath}

\def\hb{\hbox to 10.7 cm{}}

\begin{document}

\def\thepage{}

\begin{frontmatter}              

\title{Cross-Domain Generalization and Knowledge Transfer in Transformers Trained on Legal Data}


\author[A]{\fnms{Jarom\'{i}r} \snm{\v{S}avelka}\thanks{E-mail: jsavelka@andrew.cmu.edu}},
\author[B]{\fnms{Hannes} \snm{Westermann}}, and
\author[B]{\fnms{Karim} \snm{Benyekhlef}}

\runningauthor{Savelka and Westermann}
\address[A]{School of Computer Science, Carnegie Mellon University}
\address[B]{Cyberjustice Laboratory, Facult\'e de droit, Universit\'e de Montréal}

\begin{abstract}
We analyze the ability of pre-trained language models to transfer knowledge among datasets annotated with different type systems and to generalize beyond the domain and dataset they were trained on. We create a meta task, over multiple datasets focused on the prediction of rhetorical roles. Prediction of the rhetorical role a sentence plays in a case decision is an important and often studied task in AI \& Law. Typically, it requires the annotation of a large number of sentences to train a model, which can be time-consuming and expensive. Further, the application of the models is restrained to the same dataset it was trained on. We fine-tune language models and evaluate their performance across datasets, to investigate the models' ability to generalize across domains. Our results suggest that the approach could be helpful in overcoming the cold-start problem in active or interactvie learning, and shows the ability of the models to generalize across datasets and domains.
\end{abstract}

\end{frontmatter}

\section{Introduction}
In this paper we examine the ability of pre-trained language models to generalize beyond the domain and dataset they were trained on and to transfer knowledge between datasets in the legal domain. We show that datasets created with different type systems, and in different domains and jurisdictions can be utilized towards a common goal. As high-quality legal data sets are scarce, the efficient use of those that are available is of utmost importance in AI \& Law research. We explore if and how different datasets and type systems can be re-cast in such a way that a language model fine-tuned on one (source domain) could be successfully utilized on the other (target domain). We also explore if and how a language model could benefit from the availability of data from other datasets. We address these questions by evaluating performance of language models fine-tuned on three data sets of adjudicatory decisions coming from different countries and legal domains. The three data sets are annotated in terms of rhetorical roles sentences play in a decision, each of them using its own type system. We re-cast the tasks to a common type system, train models on the individual datasets and their combinations, and evaluate the models' performance.

\section{Background}
Written legal cases typically follow a certain pattern in their reasoning. For example, legal cases often contain the following sections:
\begin{enumerate}
    \item Claims of the plaintiff and responses/counter-claims of the defendant
    \item A description of factual circumstances of the case as seen by the parties and as determined by the court
    \item Legal rules applicable to the factual circumstances
    \item Application of the rules to the factual circumstances
    \item Court's conclusions and outcomes of the case
\end{enumerate}

\noindent The sections play different roles in a decision, and could carry different meaning dependent on the context in which they are read. A judge, a law student, a party to a dispute, or a legal professional could all read the different sections with different focus depending on what they need to learn. It is an indispensable skill for any professional lawyer to quickly identify and understand the different sections of a case. 

Due to its prominence the task has attracted abundant research in AI \& Law (see e.g. \cite{bhattacharya2019, walker:2019}. Many researchers have worked on automating the identification of the different sections, using sophisticated ML models, on a sentence, paragraph or other level. The automatic segmentation of cases could serve as a reading aid for actors in the legal world, by automatically visualizing the different sections. It could also serve as a way to improve legal search results, as users of legal databases could search in specific sections (such as the application of a rule) to get more relevant results. Further, it serves as an important prerequisite for future research in AI \& Law. Researchers could leverage the information about the sections to understand their relevance for, e.g., prediction of the outcome of a case or mining of reasoning patterns or stereotypical factual occurrences.

Typically, a research in prediction of rhetorical roles includes dataset creation and training and evaluation of different ML models on that dataset. The dataset typically focuses on one or several narrowly domains. The model is then tested against a test set, typically sampled randomly from the dataset. This shows the predictive capability on test data from the same distribution as the training data. 

It can be difficult to evaluate how well a model would perform if applied to other domains, and if it is able to generalize beyond the domains captured in the dataset. For example, it is possible that a model has only learned a specific vocabulary characteristic of a single domain or jurisdiction to describe factual circumstances, rather than the general idea of what a sentence describing the circumstances looks like. 

Evaluating the models on multiple datasets, however, is not straightforward. Firstly, publicly available datasets are scarce in AI \& Law. Further, the datasets that are available typically use custom type systems, making cross-dataset evaluation challenging. In this paper, we identify an example meta type system, that generalizes across three datasets. We use the meta type system to train models on data from different domains and jurisdictions and assess performance of the models across the domains.

We compare the ability of different types of ML models to generalize from one domain to another. First, as a baseline we use a Support Vector Machines classifier which learns the correlation between word and/or n-gram occurrences and a certain label. Then, we evaluate the more recently developed BERT model (bidirectional encoder representation from transformers). This model has been pre-trained on massive corpora of text, to learn a language model. It can then be further fine-tuned on a down-stream task on a specialized dataset, achieving strong results with little training data by leveraging the previously learned language representation. 

\section{Hypotheses}

By conducting the experiments described in Section \ref{sec:experimental_design}, we investigate the following hypotheses:
\begin{itemize}
    \item \emph{H1} - The ML models are able to generalize the knowledge learned in one domain and apply it successfully in another domain.
    \item \emph{H2} - The BERT model is able to leverage its understanding of language to abstract beyond the specific domain vocabulary, thereby generalizing better across the domains than the SVM model.
    \item \emph{H3} - Training the models on pooled data from multiple datasets improves performance and results in more robust models. 
\end{itemize}

\noindent Showing that the models generalize well among different domains could have important implications for the real-world applicability of the methods. Further, they could be employed in active or interactive learning settings, where annotators are presented with predictions by the model and confirm/correct these to create accurate models. This could drastically lower the number of expensive annotations that need to be performed to start the annotation of novel datasets.

\section{Related Work}
\label{sec:related_work}
The identification of rhethorical rules of sentences in legal cases has been investigated by several researchers. \cite{savelka:2018} used Conditional Random Field models with custom features to predict the rhethorical role of sentences in three distinct domains. \cite{walker:2019} trained a number of machine learning models to predict the rhethorical role of sentences in decision from the U.S. Board of Veteran Appeals and compared them to rule-based approaches. \cite{westermann2019} created an interface to easily create boolean search rules for sentence classification. \cite{bhattacharya2019} created a dataset of decisions from the Indian Supreme Court decisions from 5 domains, and trained Hierarchical BiLSTM CRF models, achieving a macro F-1 score of up to 0.82. These papers trained and evaluated the models on the same datasets. The contribution in this paper is that we investigate the capability of models trained on one dataset to generalize to \textit{other} datasets. Therefore, we re-cast three datasets (including the publicly available datasets from \cite{walker:2019} and \cite{bhattacharya2019}) into a meta type system. Then we investigate the ability and respective performance of SVM and RoBERTa models trained on a single or multiple datasets to generalize to the other datasets. To our knowledge, training models for the classification of legal texts on one dataset and evaluating the performance on other datasets has not been extensively investigated previously, and is thus a novel contribution. 

There are numerous examples of successful applications of BERT-based models on legal texts. In \cite{Condevaux2019} BERT is evaluated on classification of claim acceptance given judges' arguments. A task of retrieving related case-law similar to a case decision a user provides is tackled in \cite{Rossi2019}. The authors demonstrate the effectiveness of using BERT for this task while focusing on mitigating the constraint on document length imposed by BERT. In \cite{Chalkidis2019} BERT is evaluated as one of the approaches to predict court decision outcome given the facts of a case. BERT has been successfully used for classification of legal areas of Supreme Court judgments. \cite{Howe2019} The authors of \cite{Rabelo2019} combine BERT with simple similarity measure to tackle the challenging task of case law entailment. BERT was also used in learning-to-rank settings for retrieval of legal news \cite{Sanchez2020} and case-law sentences interpreting statutory concepts \cite{Savelka2020}. In \cite{Westermann2020}, the researchers investigate the additional fine-tuning of language models on related tasks to improve performance in the analysis of legal entailment. 



We show that models pre-trained on one dataset can to some extent perform predictions on other datasets. This could be used to bootstrap new datasets in other domains. This follows a steady line of work in AI \& Law on making annotation more effective. Westermann et al. \cite{westermann2019} proposed and assessed a method for building strong, explainable classifiers in the form of Boolean search rules. Employing an intuitive interface, the user develops Boolean rules for matching instead of annotating the individual sentences. In \cite{westermann2020b} a method for using pre-trained language models to identify semantically similar sentences suitable for annotation is proposed. \v{S}avelka and Ashley \cite{savelka2015} evaluated the effectiveness of an approach where a user labels the documents by confirming (or correcting) the prediction of a ML algorithm (interactive approach). The application of active learning has further been explored in the context of classification of statutory provisions \cite{waltl2017} and eDiscovery \cite{cormack2016,cormack2015,hogan2009}. 

\section{Experimental Design}
\label{sec:experimental_design}
In order to evaluate the ability of models to generalize beyond a single domain, we employ an experimental design consisting of several steps. First, we identify three datasets containing a categorization of sentences by rhetorical role (Section \ref{sec:data}). Then, we identify a meta type system that we can transform all the type systems into to create a task shared among the datasets (Section \ref{sec:tasks}). We then fine-tune and evaluate a pre-trained language model and a Support Vector Machine model (Section \ref{sec:models}) on different combinations of these datasets (Sections \ref{sec:models} and \ref{sec:experiments}). 

\subsection{Data}
\label{sec:data}
In this work we utilize three datasets. The first one comes from  \cite{walker:2019}. The authors analyzed 50 fact-finding decisions issued by the U.S. Board of Veterans' Appeals (``BVA") from 2013 through 2017, all arbitrarily selected cases dealing with claims by veterans for service-related post-traumatic stress disorder (PTSD). For each of the 50 BVA decisions in the PTSD dataset, the researchers extracted all sentences addressing the factual issues related to the claim for PTSD, or for a closely-related psychiatric disorder. These were tagged with the rhetorical roles \cite{walker:2017} the sentences play in the decision. These were Finding, Reasoning, Evidence, Legal Rule, and Citation. \footnote{Dataset available at \url{https://github.com/LLTLab/VetClaims-JSON}}

The second dataset also focuses on rhetorical roles of sentences. Bhattacharya et al.~\cite{bhattacharya2019} analyzed 50 opinions of the Supreme Court of India. The cases were sampled from five different domains in proportion to their frequencies (criminal, land and property, constitutional, labor and industrial, and intellectual property). The decisions were split into 9,380 sentences and manually classified into one of the seven categories according to the rhetorical roles they play in a decision. These were Facts, Ruling (lower court), Argument, Ratio, Statute, Precedent, Ruling (present court).\footnote{Dataset available at \url{github.com/Law-AI/semantic-segmentation}}

We also created a brand-new data set by scraping the case briefs from a publicly available Case Brief Summary database.\footnote{\url{http://www.casebriefsummary.com/}, site unavailable as of 2020-11-29} The case briefs were categorized in terms of the areas of regulation, such as administrative law, business law, or criminal law (11 categories in total). In total, we were able to obtain 715 unique case briefs. The case briefs are structured into a number of sections with headings. We extract the sections based on an extensive battery of regular expressions to segment the retrieved briefs into individual sections. While there were over 100 unique section heading names we were able to identify six main types to which we could map many of the different variations (e.g., all of Legal Issue, Issues, and Issue map to a single category). We applied a specialized legal case sentence boundary detection system \cite{Savelka2017} to segment the sections into sentences. This results in the dataset comprising 9,965 sentences, each with one of the six labels corresponding to the section of the brief where the sentence occurred. These were Facts, Issue, Conclusion, Procedural History, Reasoning, and Rule.


\subsection{Task}
\label{sec:tasks}
The datasets described in \ref{sec:data} use different annotation type systems and stem from different domains and decision makers. However, parts of the type systems overlap, making it possible to map certain types to a new meta type system that allows the training of models and evaluation of the trained models between datasets. 

\begin{table}[t]
    \centering
    \begin{tabular}{l|r|r|r|r}
            Label & BVA & CB & ISC & Total\\
    \hline
    Facts & 2,420 (39\%) & 4,182 (42\%) & 2,219 (19\%) & 8,821 (32\%) \\
    Non-Facts & 3,733 (61\%) & 5,783 (58\%) & 9,380 (81\%) & 18,896 (68\%) \\
    \hline
    Total & 6,153 & 9,965  & 11,599  & 27,717  \\
    \end{tabular}
    \caption{F$_1$ scores for models trained on specific dataset training pools (rows) predicting on testing datasets (columns), compared between SVM and RoBERTa. Grey cells indicate that the training data includes the target dataset.}
    \label{tab:results_exp_1_f1}
\end{table}



In order to map the source type systems into a single type system, we identify the label of the source datasets that 

The meta type system establishes a binary classification whether a sentence states factual circumstances of the case. We do not specify an explicit definition for such a sentence. Instead, we identify the best fitting types in the BVA, Case Briefs, and Indian Supreme Court datasets. BVA dataset contains sentences labeled as Evidence (2,420 out of 6,153). An Evidence sentence is defined as a sentence that primarily states the content of the testimony of a witness, states the content of documents introduced into evidence, or describes other evidence.\footnote{\url{https://github.com/LLTLab/VetClaims-JSON}} The Case Briefs dataset does not come with a definition of the individual types. However, there is a large number of sentences coming from a section commonly labeled as Facts (4,182 out of 9,965). The Indian Supreme Court dataset contains sentences labeled as Facts (2,219 out of 11,599). A Facts sentence refers to the chronology of events that led to filing the case, and how the case evolved over time in the legal system (e.g., First Information Report at a police station, filing an appeal to the Magistrate). The task is then to predict if the sentence is of the focused type or not. While the source type systems are different, and the definitions of the corresponding types are not the same, for the purposes of the training and evaluation we assume that the new type system describes the same task when applied across the three datasets (i.e., we assume that we have transformed three related tasks into a unified one).

\subsection{Dataset splits}
We randomly split all the data sets into the training, validation, and test sets using ratios of 50-25-25. The splitting for all four datasets is performed at the level of documents, i.e., all the sentences from a single document are in the same fold. We did not take the size of a document into account, meaning that the number of sentences could vary slightly between datasets. The training split of the datasets is used for training the models, the validation split is used for model selection and hyperparameter optimization, while the test set is used for final evaluation. 

\subsection{Models}
\label{sec:models}
In this work, we use the RoBERTa (a robustly optimized BERT pretraining approach) described in \cite{Liu2019} as the starting point for our experiments.\footnote{\url{github.com/pytorch/fairseq/tree/master/examples/roberta}} Out of the available models we chose to work with the smaller roberta.base model that has 125 million parameters. This choice was motivated by the ability to iterate the experiments faster when compared to working with roberta.large with 355 million parameters. RoBERTa is using the same architecture as BERT. However, the authors of \cite{Liu2019} conducted a replication study of BERT pre-training and found that BERT was significantly undertrained. They used the insights thus gained to propose a better pre-training procedure. Their modifications include longer training with bigger batches and more data, removal of the next sentence prediction objective, training on longer sequences on average (still limited to 512 tokens), and dynamic changing of the masking pattern applied to the training data. \cite{Liu2019}

As baseline, we use a Support Vector Machine (SVM) classifier. SVM constructs a hyper-plane in a high dimensional space, which is used to separate the classes from each other. As an implementation of SVM we use the scikit-learn's Support Vector Classification module.\footnote{\url{scikit-learn.org/stable/modules/generated/sklearn.svm.LinearSVC.html}} As features we use the bag of words of (1-3)-grams weighted by TF-IDF.

\subsection{Experiments}
\label{sec:experiments}
We train the models on all the different possible pools of training data. The possible pools are BVA, CB, ISC, BVA+CB, BVA+ISC, CB+ISC, and BVA+CB+ISC. Both RoBERTa and SVM are trained on each of these pools separately, and the results are presented for prediction on the test split of each dataset (Section \ref{sec:experiments}). 


In all the experiments, we fine-tune the base RoBERTa model for 10 epochs on the training splits of the selected datasets. We use the batch size of 8 which is the maximum allowed by our hardware setup (1080Ti with 11GB) given we set the length of a sequence to 512 (maximum). As optimizer we use Adam with initial learning rate set to $4e^{-5}$. We store a model's checkpoint after the end of each training epoch. The checkpoints are evaluated on the corresponding validation set. The model with the highest F$_1$ on the validation set is then selected as the one to make predictions on the test sets. The performance on the test sets is what we report in the results section.

The SVM is optimized via gridsearch over the few most important hyperparameters ($C$, class weight, number of iterations). The various hyperparameter combinations are evaluated against the validation dataset. The model with the highest F$_1$ on the validation set is then selected as the one to make predictions on the test sets. The performance on the test sets is what we report in the results section.

For evaluation we report the F1-measure (F$_1$), i.e., the traditional information retrieval measures, to evaluate performance of the trained models. Since both the tasks are binary the application of the measure is straightforward.

\begin{align*}
\footnotesize
P=\mathlarger{\sum_{i=1}^{|S|}}\frac{TP}{TP+FP} \qquad
R=\mathlarger{\sum_{i=1}^{|S|}}\frac{TP}{TP+FN} \qquad
F_1=\frac{2PR}{P+R}
\end{align*}

\noindent In the formulas, $TP$ stands for true positives, $FP$ for false positives, and $FN$ for false negatives.




\section{Results}
\label{sec:results}

\subsection{H1 - The ML models are able to generalize the knowledge learned in one domain and apply it successfully in another domain.}

The results displayed in Table \ref{tab:results_exp_1_f1} confirm the hypothesis to some extent. Despite being trained and evaluated on two different datasets, the performance of the models on the target datasets is decent. The RoBERTa model performs stronger than the SVM model across the board, which is expected as RoBERTa is much more complex model (compared to SVM) pre-trained for language understanding. Overall, the performance appears promising to mitigate the cold-start problem in active or interactive learning. 
The RoBERTa model trained on the CB dataset achieves the highest average performance across the datasets with $F_1=0.75$ on average, underperforming the dataset dedicated models by a rather small margin (difference of 0.1 for BVA, and 0.02 for ISC). This might be due to the size and thematic breadth of this dataset. The models trained on BVA and ISC also achieve strong score (0.70 on BVA and 0.64 on ISC on averages for RoBERTa).
Interestingly, the SVM trained on the BVA dataset performs the strongest, with an average performance of .68 across all the datasets. This seems to be inflated due to the high performance on the BVA dataset, however. The SVM further performs much worse when trained on ISC, which is likely due to the ISC dataset being much more inbalanced than the other target datasets. Curiously, the SVM model even performs better on the ISC dataset when trained with BVA or CB rather than ISC itself.

\begin{table}[t]
    \centering
    \begin{tabular}{l|rr|rr|rr|rr}
               & \multicolumn{2}{c|}{BVA} & \multicolumn{2}{c|}{CB} & \multicolumn{2}{c|}{ISC} & \multicolumn{2}{c}{Avg} \\
               & SVM & RoBERTa & SVM & RoBERTa & SVM & RoBERTa & SVM & RoBERTa\\
    \hline
    BVA        & \cellcolor[HTML]{C0C0C0}.92 & \cellcolor[HTML]{C0C0C0} .94   & .67 & .71   & .44 & .44 & .68 & .70  \\
    CB         & .60 & .84   & \cellcolor[HTML]{C0C0C0} .78 & \cellcolor[HTML]{C0C0C0} .83   & .50 & .57 & .63 & .75 \\
    ISC        & .11 & .67   & .19 & .65   & \cellcolor[HTML]{C0C0C0} .41 & \cellcolor[HTML]{C0C0C0} .59 & .24 & .64\\
    \end{tabular}
    \caption{F$_1$ scores for models trained on specific dataset training pools (rows) predicting on testing datasets (columns), compared between SVM and RoBERTa. Grey cells indicate that the training data includes the target dataset.}
    \label{tab:results_exp_1_f1}
\end{table}

\subsection{H2 - The BERT model is able to leverage its understanding of language to abstract beyond the specific domain vocabulary, thereby generalizing better across the domains than the SVM model.}

The RoBERTa model appears to transfer the knowledge learned on any one dataset to prediction on the other two much better than the baseline SVM model (Table \ref{tab:results_exp_1_f1}). For example, RoBERTa trained on CB performs at $F_1=0.84$ on BVA. This is a sizeable performance hit when compared to RoBERTa trained on BVA ($F_1=0.94$). However, the SVM model goes from $F_1=0.92$ to $F_1=0.60$. While it is customary for BERT models to perform better than traditional ML models, the sizeable difference of .24 in this instance might indicate that there is a qualitatively different understanding going on, as RoBERTa can understand the semantic meaning rather than the specific vocabulary used. This is very promising for the possibility of creating robust models that can generalize between domains.

\subsection{H3 - Training the models on pooled data from multiple datasets improves performance and results in more robust models.}

The first part of this hypothesis is not confirmed by our experiments. As can be seen in Table \ref{tab:results_exp_1_f1_multi} (as compared to Table \ref{tab:results_exp_1_f1}), performance does not clearly improve for models trained on multiple datasets. However, in isolated instances the performance does seem to slighlty improve when trained on additional datasets. For example, training the model on CB and ISC to predict ISC seems to slightly improve the performance of RoBERTa as compared to the model only trained on ISC (.59 $\rightarrow$ .60), while significantly improving the performance of the SVM (.41 $\rightarrow$ .53). It is possible that larger datasets could further improve the performance. 

As for the part of the hypothesis concerning robustness, the results suggest that RoBERTa is much better at handling data coming from different datasets. It appears that when the models are trained on other datasets in addition to the target dataset, the performance matches the model trained only on the target dataset. This is not the case for the SVM baseline (e.g., consider the model trained on CB+ISC data applied to CB). This is very important as it allows training of the model on multiple domains  without any degradation in performance on those tasks. The result is a model that performs well on all three datasets. Such a model would likely generalize better to other domains (not considered here) as compared to the model trained exclusively on one of the three datasets.
This robustness can be illustrated by the fact that combining data from the two non-target datasets seems to yield higher performance on the target datasets than training just on either of the non-target datasets. For example, training the RoBERTa model on both CB and ISC yields higher performance when predicting BVA (.85) than using either CB (.84) or ISC (.67) on their own.

\begin{table}[t]
    \centering
    \begin{tabular}{l|rr|rr|rr}
               & \multicolumn{2}{c|}{BVA} & \multicolumn{2}{c|}{CB} & \multicolumn{2}{c}{ISC} \\
               & SVM & RoBERTa & SVM & RoBERTa & SVM & RoBERTa \\
    \hline
    BVA+CB     & .91 & .93   & .78 & .84   & .41 & .55 \\
    BVA+ISC    & .91 & .93   & .46 & .75   & .46 & .59 \\
    CB+ISC     & .55 & .85   & .71 & .82   & .53 & .60 \\
    BVA+CB+ISC & .90 & .93   & .68 & .82   & .50 & .59 \\
    \end{tabular}
    \caption{F1-scores for models trained on specific dataset training pools (rows) predicting on testing datasets (rows), compared between SVM and RoBERTa.}
    \label{tab:results_exp_1_f1_multi}
\end{table}



\section{Discussion}

The results reveal several interesting properties of the evaluated models. First of all, they show that it is indeed possible to train a model on one domain, and use it for prediction on another domain with decent performance. This holds especially for the powerful RoBERTa model. It should be noted that the models in \ref{tab:results_exp_1_f1} have never seen any sample from the target dataset, except when the source dataset and the target dataset are the same, and still achieve strong results. This is promising for multiple reasons. It shows a clear path towards using a model trained on one dataset to bootstrap a new dataset for active or interactive learning purposes. It also shows that the tasks defined in terms of their respective type systems are related, despite being created in different contexts and jurisdictions. The relatedness enables ML models to be successfully applied across domains.

Further, several interesting observations can be made when comparing the performance of the SVM and RoBERTa model. There are several instances where the RoBERTa model significantly outperforms the SVM model. The difference in results could potentially be explained by the RoBERTa model being able to abstract beyond the vocabulary used, to grasp the meaning and semantic type of a sentence, by leveraging the language model learned during the pre-training on general language data. The SVM model, on the other hand, only has access to the language of the specific dataset used for training. Of course, further study would be required to verify this, but the difference in results is staggering, up to 0.5 in F$_1$ scores on some datasets.

Finally, the RoBERTa models trained on multiple datasets seem to retain high performance on the specific datasets they are trained on, while also improving performance on datasets they were not trained on. This is another promising result showing that the models can learn patterns even from datasets created in different contexts and jurisdictions, and use the additional data to create robust and strong decision rules.

\section{Future Work}
We observed that it is possible to train a model on one dataset, and use it for prediction on another one with decent performance. This could be utilized for bootstrapping a strong active or interactive learning model and, hence, solving the cold-start problem. For future, we plan to validate this assumption and assess the benefit of the technique to the down-stream task (i.e., the annotation supported with active learning).

In this work, we use three datasets supporting related tasks. We performed a simple transformation on those tasks to unify their label spaces. There is a plenty of space to experiment with different transformations, i.e., tasks. Also, there is a potential to explore other existing datasets looking for those that support other types of related tasks (e.g., decision outcome prediction). It might be particularly interesting to look into the options to work with documents in different languages.

Finally, we used the base version of RoBERTa for our experiments. It is worth noting that this model, nor its large sibling are considered to be the most up-to-date state-of-the-art as of the time this paper is being published. Because of the promising results it is warranted to analyze how much improvement could one get by utilizing bigger and more powerful models.

\section{Conclusion}
In this paper, we have provided an example of how datasets created in different contexts can be re-cast to support the same task, and utilized towards a common goal. We showed an advantage in performance of pre-trained language models when applied to  domains different from the one they were trained on, compared to SVM models. The pre-trained language models seem to be able to understand the underlying idea behind a task to a larger extent than SVM models. Further, training the models on several domains improved robustness and to some extent performance. Overall, these results suggest that pre-trained language models have several desirable properties when training on multiple datasets, making them an ideal tool for efficient utilization of the existing AI \& Law datasets to support future research in novel tasks.




\begin{thebibliography}{99}
\bibitem{bhattacharya2019}
Bhattacharya, P., S. Paul, K. Ghosh, S. Ghosh, and A. Wyner. ``Identification of Rhetorical Roles of Sentences in Indian Legal Judgments.'' arXiv preprint arXiv:1911.05405 (2019).

\bibitem{Chalkidis2019}
Chalkidis, Ilias, Ion Androutsopoulos, and Nikolaos Aletras. ``Neural legal judgment prediction in english." arXiv preprint arXiv:1906.02059 (2019).

\bibitem{Condevaux2019}
Condevaux, Charles, et al. ``Weakly Supervised One-Shot Classification Using Recurrent Neural Networks with Attention: Application to Claim Acceptance Detection." JURIX. 2019.

\bibitem{cormack2016}
Cormack, G., and M. Grossman. ``Scalability of continuous active learning for reliable high-recall text classification.'' In Proc. 25th ACM Int'l  Conf. on Info. \& Knowledge Management, pp. 1039-1048. 2016.

\bibitem{cormack2015}
Cormack, G., and M. Grossman. ``Autonomy and reliability of continuous active learning for technology-assisted review.'' \emph{arXiv preprint arXiv}:1504.06868 (2015).

\bibitem{Devlin2018}
Devlin, Jacob, et al. ``Bert: Pre-training of deep bidirectional transformers for language understanding.'' arXiv preprint arXiv:1810.04805 (2018).


\bibitem{hogan2009}
Hogan, C., R. Bauer, and D. Brassil. ``Human-aided computer cognition for e-discovery.'' In \emph{Proc. 12th Int'l Conf. on Artificial Intelligence and Law}, pp. 194-201. 2009.

\bibitem{Howard2018}
Howard, Jeremy, and Sebastian Ruder. ``Universal language model fine-tuning for text classification." arXiv preprint arXiv:1801.06146 (2018).

\bibitem{Howe2019}
Howe, Jerrold Soh Tsin, Lim How Khang, and Ian Ernst Chai. ``Legal area classification: a comparative study of text classifiers on Singapore supreme court judgments." arXiv:1904.06470 (2019).

\bibitem{Lan2019}
Lan, Zhenzhong, et al. ``Albert: A lite bert for self-supervised learning of language representations." arXiv preprint arXiv:1909.11942 (2019).

\bibitem{Liu2019}
Liu, Yinhan, et al. ``Roberta: A robustly optimized bert pretraining approach." arXiv preprint arXiv:1907.11692 (2019).

\bibitem{Mikolov2013distributed}
Mikolov, Tomas, et al. ``Distributed representations of words and phrases and their compositionality." Advances in neural information processing systems. 2013.

\bibitem{Mikolov2013efficient}
Mikolov, Tomas, et al. ``Efficient estimation of word representations in vector space." arXiv preprint arXiv:1301.3781 (2013).

\bibitem{Pennington2014}
Pennington, Jeffrey, Richard Socher, and Christopher D. Manning. ``Glove: Global vectors for word representation." \emph{EMNLP}. 2014.

\bibitem{Peters2017}
Peters, Matthew E., et al. ``Semi-supervised sequence tagging with bidirectional language models." arXiv preprint arXiv:1705.00108 (2017).

\bibitem{Rabelo2019}
Rabelo, Juliano, Mi-Young Kim, and Randy Goebel. ``Combining similarity and transformer methods for case law entailment." \emph{ICAIL}. 2019.

\bibitem{Radford2018}
Radford, Alec, et al. ``Improving language understanding by generative pre-training." (2018): 12.

\bibitem{Raffel2019}
Raffel, Colin, et al. ``Exploring the limits of transfer learning with a unified text-to-text transformer." arXiv preprint arXiv:1910.10683 (2019).

\bibitem{Rossi2019}
Rossi, Julien, and Evangelos Kanoulas. ``Legal Search in Case Law and Statute Law." JURIX. 2019.

\bibitem{Sanchez2020}
Sanchez, Luis, et al. ``Easing Legal News Monitoring with Learning to Rank and BERT." European Conference on Information Retrieval. Springer, Cham, 2020.

\bibitem{Savelka2020}
Savelka, Jaromir. Discovering sentences for argumentation about the meaning of statutory terms. Diss. University of Pittsburgh, 2020.

\bibitem{savelka:2018}
Savelka, J., and Kevin D. Ashley. ``Using CRF to detect different functional types of content in decisions of united states courts with example application to sentence boundary detection.'' \textit{ASAIL 2017}.

\bibitem{Savelka2017}
Savelka, J., Walker, V. R., Grabmair, M., \& Ashley, K. D. (2017). Sentence boundary detection in adjudicatory decisions in the united states. \emph{Traitement automatique des langues}, 58, 21.

\bibitem{savelka2015}
Savelka, J., G. Trivedi, and K. Ashley. ``Applying an interactive machine learning approach to statutory analysis.'' In Proc. 28th Ann. Conf. on Legal Knowledge \& Info. Systems (JURIX'15). IOS Press. 2015.

\bibitem{Vaswani2017}
Vaswani, Ashish, et al. ``Attention is all you need." Advances in neural information processing systems. 2017.

\bibitem{walker:2019}
Walker, Vern R., et al. ``Automatic Classification of Rhetorical Roles for Sentences: Comparing Rule-Based Scripts with Machine Learning." \textit{Proceedings of ASAIL 2019} (2019).

\bibitem{walker:2017}
Walker, Vern R., et al. ``Semantic types for computational legal reasoning: propositional connectives and sentence roles in the veterans' claims dataset." \textit{Proceedings of ICAIL '17}. ACM, 2017.

\bibitem{waltl2017}
Waltl, B., J. Muhr, I. Glaser, G. Bonczek, E. Scepankova, and F. Matthes. ``Classifying Legal Norms with Active Machine Learning.'' In JURIX, pp. 11-20. 2017.

\bibitem{Wang2019}
Wang, Wei, et al. ``Structbert: Incorporating language structures into pre-training for deep language understanding." arXiv preprint arXiv:1908.04577 (2019).

\bibitem{Westermann2020}
Westermann, H, J. Savelka and K. Benyekhlef. ``Paragraph Similarity Scoring and Fine-Tuned BERT for Legal Information Retrieval and Entailment." \emph{COLIEE}. 2020.

\bibitem{westermann2020b}
Westermann, H., J. Savelka, V. Walker, K. Ashley, and K. Benyekhlef. ``Sentence Embeddings and High-speed Similarity Search for Fast Computer Assisted Annotation of Legal Documents.'' In \emph{JURIX}. 2020.

\bibitem{westermann2019}
Westermann, H., J. Savelka, V. Walker, K. Ashley, and K. Benyekhlef. ``Computer-Assisted Creation of Boolean Search Rules for Text Classification in the Legal Domain.'' In \emph{JURIX}, pp. 123-132. 2019.

\end{thebibliography}
\end{document}